# ROSFD: Robust Online Streaming Fraud Detection with Resilience to Concept Drift in Data Streams

Yelleti Vivek[1*]

[1]Department of Computer Science and Engineering,
National Institute of Technology, Warangal-506004, India

vivek.yelleti@gmail.com

**Abstract**

Continuous generation of streaming data from diverse sources, such as online transactions and digital interactions, necessitates timely fraud detection. Traditional batch processing methods often struggle to capture the rapidly evolving patterns of fraudulent activities. This paper highlights the critical importance of processing streaming data for effective fraud detection. To address the inherent challenges of latency, scalability, and concept drift in streaming environments, we propose a robust online streaming fraud detection (ROSFD) framework. Our proposed framework comprises two key stages: *(i) Stage One: Offline Model Initialization.* In this initial stage, a model is built in offline settings using incremental learning principles to overcome the "cold-start" problem. *(ii) Stage Two: Real-time Model Adaptation.* In this dynamic stage, drift detection algorithms (viz.,, DDM, EDDM, and ADWIN) are employed to identify concept drift in the incoming data stream and incrementally train the model accordingly. This "train-only-when-required" strategy drastically reduces the number of retrains needed without significantly impacting the area under the receiver operating characteristic curve (AUC). Overall, ROSFD utilizing ADWIN as the drift detection method demonstrated the best performance among the employed methods. In terms of model efficacy, Adaptive Random Forest consistently outperformed other models, achieving the highest AUC in four out of five datasets.

# 1. Introduction

Every digital interaction leaves a fleeting digital footprint, resulting in the continuous generation of data. From the subtle clicks on a website to the financial transactions and the constant signals from our various electronic devices, the huge amount of information is being generated. To understand and derive timely insights from this ever-flowing information, we need to process and analyze it as and when it happens – which results in the new paradigm called streaming data analysis [1-2].

Nowhere is the need for the real-time understanding of this digital footprint more critical than in the fight against fraud. Fraudulent activities, by their nature, often attempt to look similar to the legitimate data. However, they leave potentially revealing facts in the process [30], such as (i) unauthorized access attempt flagged in system logs, (ii) suspicious sequence of financial transactions, (iii) sudden surge in social media activity, etc. The important thing is to identify the anomalous patterns within this vast and fast-growing information as they are being created, rather than analysing historical records after some irreparable damage is done. This demands the designing of robust algorithms designed to analyze these continuously generated streams.

Streaming data is commonly classified into three types [1] as follows:
- ***Data Streams***: These streams are characterized by change in number of instances between successive streams, while the number of features remain constant. (E.g. ATM fraud detection [14], anomaly detection [3], etc.)
- ***Feature Streams***: In these streams, the number of data points arriving in successive streams is constant, but the number of features differs from one stream to another. (E.g. high-resolution images from Mars [4], image segmentation [5], etc.)
- ***Combined Streams***: These streams exhibit variability in both number of data points and features across successive streams. (E.g. Log analysis [44], image segmentation [5], etc.)

Algorithms designed for processing data streams [2] must be equipped to handle concept drift, a phenomenon where the statistical properties of the data invoking either individual features or combinations thereof, evolve over time. This drift can take various forms, including abrupt changes, gradual change in patterns, or even complete reversals in the underlying data distribution. Conversely, feature streams contend with feature drift [1, 45]. Here, the number of instances remains constant in these scenarios, but the available feature set evolves over a period of time. Failure to adapt can lead to suboptimal performance, as the model either may ignore the new information or become overwhelmed by irrelevant features. In this study, we focus on handling concept drift and working with data streams.

Handling streaming data presents key challenges [46]: latency, scalability, and the necessity for incremental machine learning (ML) model training. Often, traditional machine learning (ML) struggles with evolving stream patterns, thereby requiring computationally expensive retraining from scratch. To subside this drawback, Incremental learning emerged to, thereby allowing models to adapt to new patterns and improve robustness. We noticed several extant frameworks (detailed in Section 2.1) designed for fraud detection problems, focus on capturing concept drift in evolving patterns, triggering retraining upon performance decline. However, the major drawbacks of the extant approaches are as follows:

(i) Frequent retraining based solely on performance can bias the model towards recent data.
(ii) Further, extant methods suffer from the *'cold-start'* problem [8], where initially deployed models lack sufficient data for accurate predictions, leading to high misclassification rates until enough patterns are learned to differentiate normal and anomalous behaviour.

To address the aforementioned challenges, this study introduces a resilient online streaming fraud detection (ROSFD) framework. Our approach primarily comprises two stages: **(i) Stage-I: Offline Model Initialization.** This initial phase focuses on building a foundational model using available historical data to mitigate the cold-start problem. **(ii) Stage-II: Real-time Model Adaptation.** In this dynamic phase, incoming data streams are continuously monitored for novelty using drift detection techniques (e.g., DDM, EDDM, ADWIN). Upon detecting significant novelty, the pre-trained model from Stage-I is incrementally updated, adjusting its parameters to reflect the evolving data patterns. This *'train-only-when-required'* strategy significantly reduces the computational cost and overhead associated with frequent model retraining.

The major highlights of the project are as follows:

- The study proposes a new framework specifically designed for the challenges of detecting fraud in continuous data streams.
- Proposed framework incorporates an initial offline model-building stage using historical data to provide a pre-trained model, to address cold start problem.
- Employed drift detection methods to identify novelty in incoming data streams to minimizes the computational overhead associated with continuous learning.

- By only retraining the model when novelty is detected, the proposed framework reduces the frequency of computationally expensive full retraining, leading to lower resource consumption and operational costs.

The remainder of this paper is organized as follows: Section 2 focuses on the literature review, and Section 3 discusses the proposed framework for fraud detection. Section 4 presents the dataset description and environmental setup, discusses the results. Section 5 presents the conclusion and future work. In the end, the Appendix covers a brief overview of the employed incremental classification algorithms.

## 2. Literature Survey

### 2.1 Real-time Fraud detection Works

Carcillo et al. [6] designed scalable framework for credit card fraud detection under big data frameworks and named it Scalable Real-time Fraud Finder (SCARFF). It involves two models (i) feedback classifier: random forest model is trained on previous $f$ days, and (ii) delayed classifier: it is trained on a few old transactions where the class label information is known. They employed concept drift by utilizing the sliding window approach to update the previous $f$ days and old transaction information. The cumulative detections generated by these two models are stored in the alert table to generate the alerts. In another work, Carcillo et al. [7] employed active learning strategy to select a few number of transactions from a large set of unlabeled dataset, thereby enhancing the performance of both delayed classifier and feedback classifier.

Li et al. [8] proposed live streaming fraud detection (LIFE), which constructs heterogeneous graph learning model on heterogeneous information such as transactions, users, etc. They also employed label propagation, which is used to handle the limited number of labeled fraudulent transactions for model training. They assessed the proposed methodology performance on large-scale Taobao live-streaming platform. Nguyen et al. [9] proposed example based technique over graphs, where, they initially constructed multimodal graphs. Later, by utilizing fraud sub-graph they listed out the $k$ possible fraudulent nodes by utilizing the common utility function. Further, they employed central limit theorem to identify the concept drift thereby detecting the changing in distributions.

Lebichot et al. [10] proposed continual learning based fraud detection, where the objective is to find the best model that both maximizes accuracy and minimizes catastrophic forgetting phenomen. They deployed multi-layer perceptron (MLP) in their settings and employed various weight transfer learning techniques to update the weight from one stream to another. Further, they proposed delayed feedback to quantify the catastrophic forgetting. Yousefimehr and Ghatte [11] proposed hybrid architecture wherein, (i) One class support vector machine (OCSVM) is combined with synthetic minority oversampling technique (SMOTE) and random undersampling to effective capture the distribution of fraud instances. Now, the outputs of these hybrids are passed to second stage, and (ii) in the second stage, the outputs is analyzed by two distinct models, light gradient boosting machine (LightGBM) and long short term memory (LSTM) models. They demonstrated the real-time fraud detection on European credit card dataset.

Vivek et al. [12] proposed streaming based ATM fraud detection, where they employed sliding window approach to collect ATM transactions happened over period of time. They studied the applicability of various data balancing techniques to handle data imbalance problem. They analyzed various ML models such as Naïve Bayes (NB), random forest (RF), decision tree (DT) over various ATM fraud transaction streams.

Vynokurova et al. [13] proposed a hybrid system for fraud detection in data streams. It majorly consists of two subsystems, where the first subpart focuses on anomaly detection and the interpretation of this subsystem of anomaly type. Kareem and Muhammed [14] proposed the Isolation forest for anomaly detection in data streams. Mollaoglu et al. [15] proposed fraud detection framework on consumer usage data in the field of telecommunications. They employed outlier detection mechanisms and applied over data stream data. Arya and Sastry [16] proposed predictive framework deep ensembling framework for detecting fraud detection in real-time systems (DEAL). They handled data imbalance and effectively handled the latent transaction patterns such as spending behaviour.

Deng et al. [17] proposed the usage of autoregressive model, that combine both graph neural network and transformer serial prediction of the model. This method initially builds the complex graph to automate the feature engineering to effectively capture the potential connections between transactions. This graph is served analyzed and the fraud score is computed which is used to discriminate whether the transaction is fraudulent or not. The performance of this model is assessed over the online payments of China.

Banirostam et al. [18] proposed streaming fraud detection framework to create a profile of the cardholder and utilize the degree of the deviation of the cardholder behaviour pattern to identify the fraudulent transactions. This entire process is executed over Map/Reduce approach for parallel execution alongside the human observer also deployed to cross-verify the status.

Dai et al. [19] proposed streaming detection framework with four layers: distributed storage layer: to store the incoming streams, batch training layer: where the models are trained by invoking map reduce operations, key-value sharing layer: where the key value sharing is completed to determine the aggregated model, and streaming detection layer: where the trained model is deployed to detect the anomaly scores.

In addition to the above works, several fraud detection frameworks are proposed under streaming settings such as phishing fraud detection [20], incremental fraud detection framework over evolving graphs [21], real-time credit card fraud detection under spark [22], etc.

## 2.2 Static Fraud detection Works

In this sub-section, we covered latest fraud detection works under Zeng et al. [23] proposed neural network based ensemble learning with generation (NNEnsLeG) in e-commerce payment fraud detection. They handled data imbalance by utilizing generative adversarial network (GAN) and the MLP is used as baseline classifier by utilizing bagging and boosting techniques.

Ranganatha and Mustafa [24] proposed 3D quasi recurrent neural network and proof of voting consensus blockchain technologies (Bi-3DQRNN-PoV-FD-MT) to improve the efficiency of fraud detection in mobile transactions. They employed self adaptive over sampling technique (SASOS) to balance the data and later Bi-3DQRNN is employed as the classifier to determine whether the transaction is fraudulent or non-fraudulent. Further, intelligent enhanced artificial gorilla troops (EAGTO) is proposed to tune the weights of the model parameters. In the end, the fraudulent transactions are forecasted by using PoV.

Kapadiya et al. [25] proposed ensemble learning based approach where they utilized blockchain security to identify fraudulent claims in healthcare insurance claim fraud detection. They integrated the data from three different sources, (i) in-patient data, (ii) out-patient data, and (iii) beneficiary data.  various forms of patient data. While predicting they employed ensemble model strategy to identify the fraudulent claims. Habibpour et al. [26] quantified the uncertainty to identify the change in data distributions in credit card transactions over a period of time. Here, they proposed ensemble monte carlo dropout for credit card fraud detection. This method is particularly useful in mitigating model unfairness and also allows practitioners to develop trustworthy systems.

Aftabi et al. [27] proposed GAN and ensemble based fraud detection in annual financial statements which was collected from 10 Iranian banks. The proposed approach consists of three components, (i) outlier generation: where the outliers are generated by using original data and noise variable, (ii) both real samples and generated outliers are combined and labeled, and (iii) an ensemble model is employed to classify the samples into fraudulent or non-fraudulent class.

Priyanka et al. [28] performed one class classification benchmark study on various fraud detection problems pertained to banking and cybsecurity domains. In the training phase, negative class data is initially subjects the negative class to K-Means clustering algorithm, where the optimal cluster centroids are estimated with silhouette score. Later, the thresholds are obtained based on the maximum intra-class centroid distance. During the test phase, positive class is discriminated by using these thresholds. Vivek et al. [29] proposed OCC method where during the training phase, negative class sampled data is subjected to K-Means. Later, the rules are generated from each cluster where the rule comprises lower and upper bounds of feature. In another work, Vivek et al. [30] proposed explainable and causal inference based ATM fraud detection.

Our proposed methodology is distinct from the extant works in the following ways:

- Even though some measures are employed to identify the concept drift, the extant works discussed in Section 2.1 did not address the concept drift and incorporated the findings while training the model.
- None of the extant works discussed works in Section 2.2 are not suitable to handle data streams.

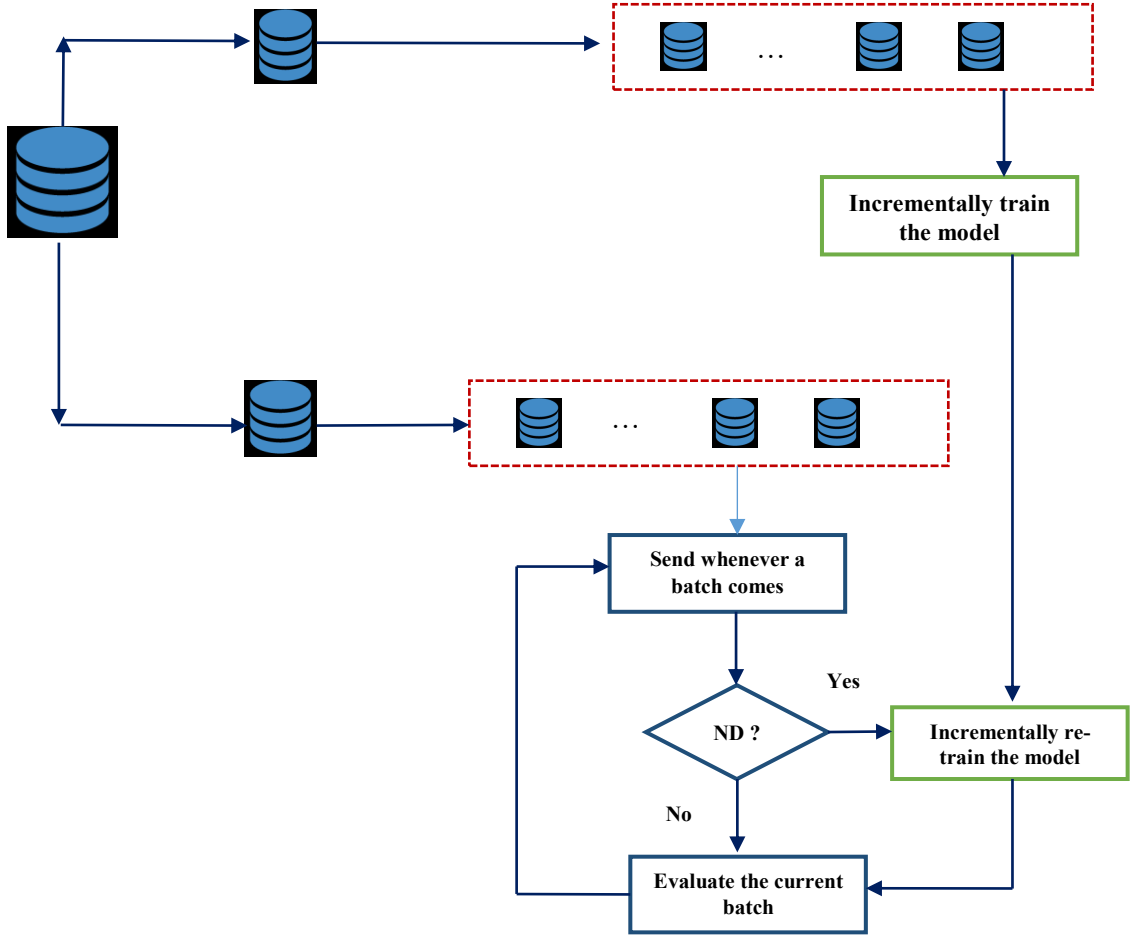

**Fig. 1. Flowchart for ROSFD algorithm**

## 3. Proposed Methodology

Our proposed methodology is presented in a step-wise manner in Algorithm 1, and the corresponding flowchart is depicted in Fig. 1. The proposed methodology is named as robust online streaming fraud detection (ROSFD) consists of the following stages:

(i) **Stage-I:** *Offline model building* initially, historical data is collected over a period of time and split into training (k%) and test sets (1-k%). Now, the machine learning (ML) model such as random forest, decision tree, batch incremental classifier etc. (see Appendix for complete details), are selected. The chosen model is then trained incrementally on the training data. It is worthwhile to note that the entire process is performed under static settings.

(ii) **Stage-II:** *Real-time Model building* The test dataset collected is partitioned into multiple sequential batches, which are processed one after another. Each incoming stream is analyzed to detect the novelty using the employed drift detection measures such as drift detection method (DDM), early drift detection method (EDDM), adaptive drift windowing method for drift detection (ADWM). Upon identifying the the novelty information, the pre-trained model obtained from Stage-I is trained in an incremental fashion and the parameters are adjusted accordingly. In this way, "train-only-when-required" strategy minimizes the burden over number of retrains thereby reducing the cost. After each update, the incoming stream is evaluated and this process is iterated until all the streams are processed.

**Algorithm 1: ROSFD algorithm**

**Input:** $M_0$: Initial model trained in offline-settings, D: Data collected
**Output:** SENS: Sensitivity score, SPEC: Specificity score, AUC: AUC score, $M_k$: Model trained for *k* times

*// Stage-I Offline model building*
1. D ← Collect the required number of samples for the local training
2. TrainD, TestD ← Train_Test_Split(D)
2. batches ← ConvertIntoBatches(TrainD)
3. $M_0$ ← Initialize the model
4. k ← getLength(batches)
5. SENS, SPEC, AUC ← []
6. **while** i < k **do**
7.     $M_i$ ← IncrementallyTrain($M_i$, k)

*// Stage-II Real-time Model building on TestD*
8. **while** new batch **do**
9.     flag ← IdentifyConceptDrift(new batch) // Eq. 1-3
10.     **if** flag = = True **do**
11.         $M_{k+1}$ ← IncrementallyTrain($M_k$)
12.         k ← k+1
13.         pred ← Predict($M_k$, new batch)
14.         sens, spec, auc ← ComputeScores(pred,true)
15.     **else**
16.         pred ← Predict($M_k$, new batch)
17.         sens, spec, auc ← ComputeScores(pred,true)
18.     append (SENS, sens)
19.     append (SPEC, spec)
20.     append (AUC, auc)
21. **return** SENS, SPEC, AUC

**Stage-I: Offline model building**

This stage focuses on training a baseline model using the available historical data before transitioning to real-time learning.

1. **Step-1:** *Data Collection* The foremost step of this framework is to collect the historical data over a period of time. Now, the collected dataset is split into training (k%) and test datasets (1-k%) by following stratified random sampling, thereby maintaining the same proportions of both negative and positive classes ratio in training and test sets.

2. **Step-2:** *Choose a ML model* In this step, one of machine learning (ML) models is chosen such as random forest, decision tree, batch incremental classifier etc. Thereafter the model is initialized with the respective parameters.

3. **Step-3:** *Incremental Model building under static settings*
   a. In this step, the chosen model from step-2 is built over the training dataset of step-1.
   b. However, instead of training the model at once with the entire training dataset, we chosen the incremental way of building the model.
   c. This is achieved by partitioning the training data into small partitions and the model is updated progressively with small partitions of the training data.
   d. By adopting this approach ensures that the model can effectively handle sudden drifts in the data distribution within the incoming stream.

After following the above three steps, the trained model is deployed in real-time settings, where it is continuously adapted to new information by following an incremental learning approach.

**Stage-II: Real-time model building**

It is worthwhile to note that, Stage-I is completely operated under static settings. Now, in Stage-II, we operate completely under streaming settings. The test dataset from stage-I is utilized here in Stage-II. This test dataset is divided into multiple stream batches and analyzed one after another, where the model is incremental adopted and trained only when the novelty is detected, i.e., only when the new information is identified. This strategy helps the model to stay up-to-date and robust while handling complex scenarios.

4. **Step-4:** *Collecting stream* Now, we are operating in streaming settings, the stream is sent over to the deployed real-time system.

5. **Step-5:** *Invoking novelty detection method*
   a. In this step, the received stream is subjected to detect the novelty by utilizing one of the three data drift detection methods, i.e., DDM, EDDM or ADWM.
   b. **DDM:** Drift detection method (DDM) [31] is a concept drift detection technique based on the probably approximately correct (PAC) learning model. It assumes that, in a stationary environment, the error rate of a learner decreases as more samples are analyzed. However, if the error rate increase than the pre-specified threshold, then it indicates the detection of novelty in the incoming stream.

   c. **EDDM:** Early drift detection method (EDDM) [32] is also similar to DDM, however instead of complete error rate, it tracks average distance between errors. If the distance between model errors is too far, then it concludes that the incoming stream has the novelty information.
   d. **ADWIN:** Adaptive winnowing (ADWIN) [33] method detects the novelty in the data distribution by adjusting the size of the data window by following window based approach. It keeps track of the data distribution and upon identifying any huge variation indicates the novelty in the incoming stream.
   e. If the novelty is detected by any of the above-discussed methods, then Step 6, followed by Step 7 are invoked. Otherwise, Step 7 is invoked.
6. **Step-6:** *Incremental Training* Since the new information arrives, we incrementally train the model and then adjust the model parameters accordingly. In this way, the continuous learning paradigm is introduced in the streaming framework. Further, the error rate is decreased, thereby improving the accuracy of the model.
7. **Step-7:** *Predictions* Now, the final prediction over the incoming stream is predicted by the model (either by step 6 or the last updated model).

The above steps (Step 4-7) are repeated until all the streams are processed completely. Later, the results of the model, i.e., sensitivity, specificity, and AUC are reported.

**Table 1: Description of the datasets**

| Dataset | #Samples | #Features | Class distribution |
|---|---|---|---|
| | | | **Negative: Positive** |
| **Credit card Churn prediction dataset [48]** | 10,000 | 13 | 80:20 |
| **Credit Card fraud Dataset [49]** | 2,94,000 | 30 | 99.94:0.06 |
| **Ethereum fraud detection dataset [49]** | 9,841 | 51 | 86.3:13.7 |
| **Distributed Denial of Service Attack (DDoS) on Software defined network dataset [49]** | 1,04,345 | 26 | 63:37 |
| **Phishing detection dataset [49]** | 10,000 | 50 | 50:50 |

# 4. Results & Discussion

## 4.1 Environmental Setup

The experimental setup for this study involved an i5 8th generation processor with eight cores running at 2.4 GHz. All experiments were conducted using Python version 3.8 within an Anaconda environment, utilizing the Jupyter Notebook.

## 4.2 Dataset description

We evaluated our approach using seven established benchmark datasets, whose key characteristics are summarized in Table 1. These datasets represent two distinct application domains: four from finance and three from healthcare. Notably, all datasets are designed for binary classification tasks, and four out of five datasets exhibit a high degree of class imbalance. We established a consistent data partitioning strategy, allocating 30% of the data to represent the historical dataset used for initial offline training of the machine learning models under static conditions, while the remaining 70% was reserved to simulate the continuous, real-time streaming environment.

## 4.3 Evaluation Metrics

Area Under the Receiver Operating Characteristic Curve (AUC) was selected as the primary evaluation metric due to its robustness, particularly when dealing with imbalanced datasets. AUC represents the probability that a randomly chosen positive instance will be ranked higher than a randomly chosen negative instance by the classifier. Mathematically, it can be defined as the average of the model's sensitivity (True Positive Rate) and specificity (True Negative Rate), effectively summarizing the trade-off between these two measures across various classification thresholds.

$$AUC = \frac{(Sensitivity + Specificity)}{2} \quad (1)$$

Where,

$$Sensitivity = \frac{TP}{TP+FN} \text{ and} \quad (2)$$

$$Specificity = \frac{TN}{TN+FP} \quad (3)$$

where TP is a true positive, FN is a false negative, TN is a true negative, and FP is a false positive.

## 4.4 Comparative analysis over AUC

In this subsection, we demonstrated the efficacy of incorporating a data drift detection method to achieve a higher mean AUC across various incremental classifiers, such as NB, KNN, ARF, and VFDT. The results for all benchmark datasets (see Table 1) are presented in Tables 2-6, with the corresponding number of retrains per strategy depicted in Figures 2-6 and the execution time in Figures 7-11. To assess the benefit of incorporating novelty detection (ND) or data drift detection methods, we conducted experiments using the following strategies:

- **Strategy 1: Without ND measures:** This framework follows Stage-I of the proposed methodology (see Section 3). However, no ND measures are incorporated in Stage-II. Consequently, retraining is triggered automatically when the AUC performance declines.
- **Strategy 2: ROSFD + DDM:** This framework follows the proposed methodology, employing DDM as the ND metric in Stage-II.
- **Strategy 3: ROSFD + EDDM:** This framework follows the proposed methodology, employing EDDM as the ND metric in Stage-II.
- **Strategy 4: ROSFD + ADWM:** This framework follows the proposed methodology, employing ADWM as the ND metric in Stage-II.

The experimental findings, presented in Tables 2 through 6, demonstrate significant performance improvements in terms of mean AUC achieved after employing ADWM as the ND metric within the proposed framework (Strategy 4). This highlights the necessity of dynamic model adaptation in response to the evolving characteristics of data streams. By initiating model updates upon the statistical identification of significant deviations from established data distributions, ADWM enables the majority of the evaluated algorithms to maintain and often improve their capability for accurate predictions, thereby identifying fraudulent activities effectively. However, we noticed some exceptions, such as the decrease in mean AUC for VFDT and HAT on the Credit Card dataset (Table 3) with ADWM. This suggests that the inherent inductive biases of certain tree-based model architectures may render them susceptible to overfitting or model instability when subjected to the potentially higher frequency of updates elicited by ADWM's adaptive windowing mechanism. In these specific instances, ADWM led the models to prioritize transient noise or inconsequential drifts, thereby impeding their ability to generalize to more fundamental, long-term fraudulent patterns inherent within that particular dataset.

Overall, the superior performance achieved by the models under Strategy 4, utilizing ADWM as the most effective ND strategy, emphasizes the critical equilibrium required in novelty detection: achieving a high degree of responsiveness to emergent changes without compromising the inherent stability of the learned model representations. ADWM's adaptive window size likely balanced both AUC and adaptation to abrupt concept drifts more effectively compared to the fixed-window paradigms underpinning DDM and EDDM. This enhanced mechanism facilitates more timely model recalibration, enabling the algorithms to assimilate newly emerging fraudulent tactics into their discriminative boundaries.

The inherent trade-off, evidenced by the elevated retraining frequency associated with ADWM, lies in the potential for model over-adaptation. While more frequent updates generally equip models to remain current with evolving data dynamics, it concurrently elevates the risk of the model becoming unduly influenced by minor, non-critical fluctuations within the data stream. The anomalous behavior of VFDT and HAT on the Credit Card dataset underscores the proposition that the optimal ND strategy may be intrinsically model-dependent and contingent upon the specific characteristics of the underlying data, including the nature and velocity of concept drift.

Now, we will discuss the individual model performances across the five benchmark datasets, explaining the distinct strengths and limitations in their aptitude for addressing the challenges of fraud detection in various datasets:

- **Adaptive Random Forest (ARF):** ARF consistently outperformed other models, attaining the highest Area Under the Receiver Operating Characteristic Curve (AUC) across multiple datasets (e.g., Churn Prediction, DDoS on SDN, and Phishing Detection). Its ensemble of decision trees, coupled with integrated mechanisms for accommodating drift and feature evolution, appears particularly well-suited for the intricate and temporally dynamic nature of fraudulent activities. The consistently positive influence of ADWM on ARF further substantiates its capacity to effectively leverage timely model updates.
- **Batch Incremental Classifier (BIC):** BIC also demonstrated robust and stable performance, often exhibiting significant performance gains when coupled with ADWM-triggered updates. Its batch-based incremental learning methodology appears effective in capturing evolving trends within the data without exhibiting excessive sensitivity to individual, potentially noisy, instances.

- **Tree-Based Models (VFDT, HT, HAT):** While generally benefiting from the incorporation of ND metrics, tree-based models exhibited less consistent performance in terms of AUC. The performance degradation observed for VFDT and HAT with ADWM on the Credit Card dataset highlights the critical importance of considering the inherent inductive biases of the learning algorithm in relation to the granularity and frequency of updates dictated by the drift detection strategy.

- **K-Nearest Neighbors (KNN):** The comparatively lower overall performance of KNN likely stems from its inherent computational complexity in high-dimensional streaming data.

- **Ensemble Methods (AWE, DWMC, AEEC):** The ensemble-based methods (Accuracy Weighted Ensemble (AWE), Dynamic Weighted Majority Classifier (DWMC), and Adaptive Expert Ensemble Classifier (AEEC)) yielded a more heterogeneous spectrum of results. None of these models consistently ranked within the top three positions across different benchmark datasets. Their efficacy appears to be heavily reliant on the diversity and individual learning capacities of their constituent classifiers, as well as the effectiveness of their weighting and adaptation mechanisms in the context of evolving fraudulent patterns.

- **Naive Bayes (NB):** NB consistently provided a relatively stable, albeit generally not superior, level of performance across the various datasets. The main drawback of this model, its stringent assumption of feature independence, likely constrains its ability to effectively model the complex and temporally evolving inter-feature relationships that often characterize sophisticated fraudulent activities.

**Table 2: Results obtained by various models on Churn Prediction dataset**

| Model | Without ND measures | | | ROSFD + DDM | | | ROSFD + EDDM | | | ROSFD + ADWM | | |
|---|---|---|---|---|---|---|---|---|---|---|---|---|
| | Sensitivity ↑ | Specificity ↑ | AUC↑ | Sensitivity ↑ | Specificity ↑ | AUC ↑ | Sensitivity ↑ | Specificity ↑ | AUC↑ | Sensitivity ↑ | Specificity ↑ | AUC↑ |
| Naive Bayes (NB) | 0.545 | 0.857 | 0.701 | 0.545 | 0.857 | 0.701 | 0.539 | 0.857 | 0.698 | 0.554 | 0.857 | 0.705 |
| K-Nearest Neighbour (KNN) | 0.239 | 0.801 | 0.520 | 0.239 | 0.801 | 0.520 | 0.226 | 0.800 | 0.513 | 0.508 | 0.820 | 0.664 |
| Adaptive Random Forest (ARF) | 0.794 | 0.843 | 0.819 | 0.800 | 0.848 | 0.824 | 0.829 | 0.844 | 0.833 | **0.928** | **0.915** | **0.920** |
| Very Fast Decision Tree(VFDT) | 0.545 | 0.857 | 0.701 | 0.545 | 0.857 | 0.701 | 0.539 | 0.857 | 0.698 | 0.569 | 0.867 | 0.718 |
| Hoeffdding Tree (HT) | 0.545 | 0.857 | 0.701 | 0.545 | 0.857 | 0.701 | 0.539 | 0.857 | 0.698 | 0.669 | 0.861 | 0.780 |
| Hoeffdding Adaptive Tree (HAT) | 0.688 | 0.848 | 0.768 | 0.700 | 0.848 | 0.774 | 0.667 | 0.851 | 0.759 | 0.708 | 0.860 | 0.784 |
| Accuracy Weighted Ensemble (AWE) | 0.0 | 0.798 | 0.399 | 0.0 | 0.798 | 0.399 | 0.0 | 0.798 | 0.399 | 0.0 | 0.798 | 0.399 |
| Batch Incremental Classifier (BIC) | 0.752 | 0.856 | 0.804 | 0.742 | 0.858 | 0.799 | 0.728 | 0.860 | 0.794 | 0.799 | 0.864 | 0.831 |
| Dynamic Weighted Majority Classifier (DWMC) | 0.570 | 0.855 | 0.713 | 0.570 | 0.856 | 0.713 | 0.550 | 0.860 | 0.705 | 0.621 | 0.853 | 0.737 |
| Adaptive Expert Ensemble Classifier (AEEC) | 0.545 | 0.857 | 0.701 | 0.545 | 0.857 | 0.701 | 0.539 | 0.857 | 0.698 | 0.554 | 0.857 | 0.705 |

*where ↑ denotes the metric should be higher

**Table 3: Results obtained by various models on Credit card dataset**

| Model | Without CD measures | | | ROSFD + DDM | | | ROSFD + EDDM | | | ROSFD + ADWM | | |
|---|---|---|---|---|---|---|---|---|---|---|---|---|
| | Sensitivity ↑ | Specificity ↑ | AUC↑ | Sensitivity ↑ | Specificity ↑ | AUC ↑ | Sensitivity ↑ | Specificity ↑ | AUC↑ | Sensitivity ↑ | Specificity ↑ | AUC↑ |
| Naive Bayes (NB) | 0.327 | 0.884 | 0.605 | 0.326 | 0.884 | 0.605 | 0.333 | 0.885 | 0.609 | 0.384 | 0.880 | 0.632 |
| K-Nearest Neighbour (KNN) | 0.334 | 0.793 | 0.563 | 0.312 | 0.795 | 0.554 | 0.338 | 0.800 | 0.569 | 0.378 | 0.798 | 0.588 |
| Adaptive Random Forest (ARF) | 0.639 | 0.835 | 0.737 | 0.638 | 0.831 | 0.734 | 0.655 | 0.837 | 0.746 | **0.688** | **0.849** | **0.769** |
| Very Fast Decision Tree(VFDT) | 0.290 | 0.869 | 0.579 | 0.290 | 0.869 | 0.579 | 0.298 | 0.873 | 0.586 | 0.237 | 0.811 | 0.524 |
| Hoeffdding Tree (HT) | 0.691 | 0.835 | 0.763 | 0.691 | 0.835 | 0.763 | 0.691 | 0.835 | 0.763 | 0.691 | 0.835 | 0.763 |
| Hoeffdding Adaptive Tree (HAT) | 0.0 | 0.781 | 0.390 | 0.0 | 0.781 | 0.390 | 0.046 | 0.618 | 0.332 | 0.039 | 0.640 | 0.340 |
| Accuracy Weighted Ensemble (AWE) | 0.657 | 0.845 | 0.751 | 0.657 | 0.844 | 0.750 | 0.659 | 0.843 | 0.751 | 0.661 | 0.844 | 0.753 |
| Batch Incremental Classifier (BIC) | 0.320 | 0.881 | 0.600 | 0.338 | 0.877 | 0.608 | 0.382 | 0.874 | 0.628 | 0.403 | 0.867 | 0.635 |
| Dynamic Weighted Majority Classifier (DWMC) | 0.327 | 0.884 | 0.605 | 0.326 | 0.884 | 0.605 | 0.333 | 0.885 | 0.609 | 0.384 | 0.880 | 0.632 |
| Adaptive Expert Ensemble Classifier (AEEC) | 0.327 | 0.884 | 0.605 | 0.326 | 0.884 | 0.605 | 0.333 | 0.885 | 0.609 | 0.384 | 0.880 | 0.632 |

*where ↑ denotes the metric should be higher

**Table 4: Results obtained by various models for DDoS on SDN dataset**

| Model | Without CD measures | | | ROSFD + DDM | | | ROSFD + EDDM | | | ROSFD + ADWM | | |
|---|---|---|---|---|---|---|---|---|---|---|---|---|
| | Sensitivity ↑ | Specificity ↑ | AUC↑ | Sensitivity ↑ | Specificity ↑ | AUC ↑ | Sensitivity ↑ | Specificity ↑ | AUC↑ | Sensitivity ↑ | Specificity ↑ | AUC↑ |
| Naive Bayes (NB) | 0.540 | 0.727 | 0.633 | 0.541 | 0.727 | 0.634 | 0.540 | 0.720 | 0.633 | 0.539 | 0.725 | 0.632 |
| K-Nearest Neighbour (KNN) | 0.590 | 0.728 | 0.659 | 0.601 | 0.732 | 0.667 | 0.600 | 0.725 | 0.663 | 0.620 | 0.736 | 0.678 |
| Adaptive Random Forest (ARF) | 0.969 | 0.993 | 0.981 | 0.964 | 0.992 | 0.978 | 0.970 | 0.993 | 0.982 | **0.972** | **0.990** | **0.981** |
| Very Fast Decision Tree(VFDT) | 0.949 | 0.9355 | 0.942 | 0.949 | 0.9355 | 0.942 | 0.940 | 0.933 | 0.937 | 0.958 | 0.933 | 0.946 |
| Hoeffdding Tree (HT) | 0.030 | 0.574 | 0.302 | 0.002 | 0.607 | 0.304 | 0.006 | 0.609 | 0.305 | 0.003 | 0.605 | 0.304 |
| Hoeffdding Adaptive Tree (HAT) | 0.030 | 0.574 | 0.302 | 0.002 | 0.607 | 0.304 | 0.006 | 0.609 | 0.305 | 0.003 | 0.605 | 0.304 |
| Accuracy Weighted Ensemble (AWE) | 0.938 | 0.986 | 0.962 | 0.937 | 0.986 | 0.961 | 0.938 | 0.986 | 0.962 | 0.941 | 0.984 | 0.962 |
| Batch Incremental Classifier (BIC) | 0.534 | 0.719 | 0.626 | 0.556 | 0.752 | 0.654 | 0.490 | 0.688 | 0.589 | 0.542 | 0.727 | 0.635 |
| Dynamic Weighted Majority Classifier (DWMC) | 0.540 | 0.727 | 0.633 | 0.541 | 0.727 | 0.634 | 0.540 | 0.726 | 0.633 | 0.539 | 0.725 | 0.632 |
| Adaptive Expert Ensemble Classifier (AEEC) | 0.540 | 0.727 | 0.633 | 0.541 | 0.727 | 0.634 | 0.540 | 0.720 | 0.633 | 0.539 | 0.725 | 0.632 |

*where ↑ denotes the metric should be higher

**Table 5: Results obtained by various models on Ethereum fraud detection dataset**

| Model | Without CD measures | | | ROSFD + DDM | | | ROSFD + EDDM | | | ROSFD + ADWM | | |
|---|---|---|---|---|---|---|---|---|---|---|---|---|
| | Sensitivity ↑ | Specificity ↑ | AUC↑ | Sensitivity ↑ | Specificity ↑ | AUC ↑ | Sensitivity ↑ | Specificity ↑ | AUC↑ | Sensitivity ↑ | Specificity ↑ | AUC↑ |
| Naive Bayes (NB) | 0.988 | 0.974 | 0.981 | 0.988 | 0.974 | 0.981 | 0.993 | 0.988 | 0.990 | 0.993 | 0.987 | 0.990 |
| K-Nearest Neighbour (KNN) | 0.662 | 0.925 | 0.794 | 0.656 | 0.925 | 0.790 | 0.771 | 0.938 | 0.855 | 0.761 | 0.936 | 0.849 |
| Adaptive Random Forest (ARF) | 0.968 | 0.975 | 0.971 | 0.992 | 0.970 | 0.981 | 1.0 | 0.991 | 0.995 | **1.0** | **0.991** | **0.995** |
| Very Fast Decision Tree(VFDT) | 0.988 | 0.972 | 0.980 | 0.988 | 0.972 | 0.980 | 0.995 | 0.995 | 0.993 | 0.989 | 0.994 | 0.992 |
| Hoeffdding Tree (HT) | 0.988 | 0.972 | 0.980 | 0.988 | 0.972 | 0.980 | 0.996 | 0.989 | 0.993 | 0.994 | 0.988 | 0.991 |
| Hoeffdding Adaptive Tree (HAT) | 0.0 | 0.850 | 0.425 | 0.0 | 0.850 | 0.425 | 0.0 | 0.850 | 0.425 | 0.0 | 0.855 | 0.425 |
| Accuracy Weighted Ensemble (AWE) | 1.0 | 0.979 | 0.989 | 1.0 | 0.979 | 0.989 | 1.0 | 0.979 | 0.989 | 1.0 | 0.979 | 0.989 |
| Batch Incremental Classifier (BIC) | 0.988 | 0.974 | 0.981 | 0.988 | 0.974 | 0.981 | 0.99w | 0.991 | 0.992 | 0.992 | 0.990 | 0.991 |
| Dynamic Weighted Majority Classifier (DWMC) | 0.988 | 0.974 | 0.981 | 0.988 | 0.974 | 0.981 | 0.993 | 0.988 | 0.990 | 0.993 | 0.987 | 0.990 |
| Adaptive Expert Ensemble Classifier (AEEC) | 0.988 | 0.974 | 0.981 | 0.988 | 0.974 | 0.981 | 0.993 | 0.988 | 0.990 | 0.993 | 0.987 | 0.990 |

*where ↑ denotes the metric should be higher

**Table 6: Results obtained by various models on Phishing detection dataset**

| Model | Without CD measures | | | ROSFD + DDM | | | ROSFD + EDDM | | | ROSFD + ADWM | | |
|---|---|---|---|---|---|---|---|---|---|---|---|---|
| | sensitivity | specifivity | AUC | sensitivity | specifivity | AUC | sensitivity | specifivity | AUC | sensitivity | specifivity | AUC |
| Naive Bayes (NB) | 0.908 | 0.808 | 0.858 | 0.908 | 0.808 | 0.858 | 0.910 | 0.830 | 0.870 | 0.910 | 0.829 | 0.870 |
| K-Nearest Neighbour (KNN) | 0.788 | 0.815 | 0.801 | 0.788 | 0.815 | 0.801 | 0.846 | 0.878 | 0.862 | 0.841 | 0.872 | 0.857 |
| Adaptive Random Forest (ARF) | 0.977 | 0.923 | 0.950 | 0.965 | 0.929 | 0.947 | 0.991 | 0.990 | 0.990 | **0.993** | **0.988** | **0.990** |
| Very Fast Decision Tree(VFDT) | 0.913 | 0.562 | 0.738 | 0.913 | 0.562 | 0.738 | 0.938 | 0.917 | 0.928 | 0.942 | 0.898 | 0.920 |
| Hoeffdding Tree (HT) | 0.913 | 0.562 | 0.738 | 0.913 | 0.562 | 0.738 | 0.913 | 0.927 | 0.920 | 0.919 | 0.907 | 0.913 |
| Hoeffdding Adaptive Tree (HAT) | 0.458 | 0.054 | 0.256 | 0.458 | 0.054 | 0.256 | 0.419 | 0.10 | 0.259 | 0.405 | 0.115 | 0.260 |
| Accuracy Weighted Ensemble (AWE) | 0.934 | 0.943 | 0.939 | 0.939 | 0.943 | 0.941 | 0.927 | 0.954 | 0.941 | 0.929 | 0.958 | 0.943 |
| Batch Incremental Classifier (BIC) | 0.907 | 0.809 | 0.858 | 0.907 | 0.809 | 0.858 | 0.910 | 0.830 | 0.870 | 0.913 | 0.827 | 0.870 |
| Dynamic Weighted Majority Classifier (DWMC) | 0.908 | 0.808 | 0.858 | 0.908 | 0.808 | 0.858 | 0.910 | 0.830 | 0.870 | 0.910 | 0.829 | 0.870 |
| Adaptive Expert Ensemble Classifier (AEEC) | 0.908 | 0.808 | 0.858 | 0.908 | 0.808 | 0.858 | 0.910 | 0.830 | 0.870 | 0.910 | 0.829 | 0.870 |

*where ↑ denotes the metric should be higher

## 4.5 Cost Benefit analysis

It is important to acknowledge that determining the best-performing model solely based on the AUC is inadequate for practical deployment. Hence, we conducted a comprehensive cost-benefit analysis, considering the following critical metrics: (i) higher mean AUC; (ii) a lower number of model retrains; and (iii) minimal computational execution time.

Our observations indicate that while ARF often exhibits superior performance compared to other models, a more informed decision-making process is necessary. Specifically, if the performance difference between ARF and the next best performing model is marginal (mean AUC difference of ≤ 1%), preference is accorded to the model with lower computational execution time. This prioritizes efficiency without significant degradation in classifier performance.

To facilitate a thorough evaluation, the aforementioned cost-benefit analysis was conducted through two distinct approaches: (i) assessing the trade-offs between different novelty detection strategies with respect to the ARF model's performance, and (ii) comparing the top two performing models within the most effective ND strategy, considering their respective AUC, retraining costs, and computational overhead.

Initially, we compared the performance across different strategies. The following are the important observations made:

- In the churn prediction and credit card fraud detection datasets, ARF with ADWM emerged as the best strategy, demonstrating a significant improvement in AUC (see Tables 2-3).
- In the SDN dataset, ARF without any ND measures and ARF with ADWM exhibited similar mean AUC. However, ADWM achieved this comparable AUC with over 100 retrains (see Fig. 5). Therefore, ARF without any ND measures is deemed the better choice due to its lower complexity and fewer retrains.
- Similarly, in the Ethereum fraud detection and phishing datasets, ARF with EDDM and ARF with ADWM showed similar mean AUC. Here, the EDDM strategy was identified as superior due to its significantly lower number of retrains (see Fig. 4 and Fig. 6).

Now, regarding the comparison of different models across different scenarios, the following are the important observations:

- In the churn prediction dataset, BIC proved to be the most advantageous model due to its very minimal execution time.
- In the credit card fraud detection dataset, HT emerged as the optimal choice, characterized by its very minimal execution time.
- Despite ARF's slightly higher execution time compared to AWE, ARF remained the preferred model in the SDN dataset owing to its superior mean AUC.
- The VFDT model outperformed the ARF model in both the Ethereum fraud detection and phishing datasets.

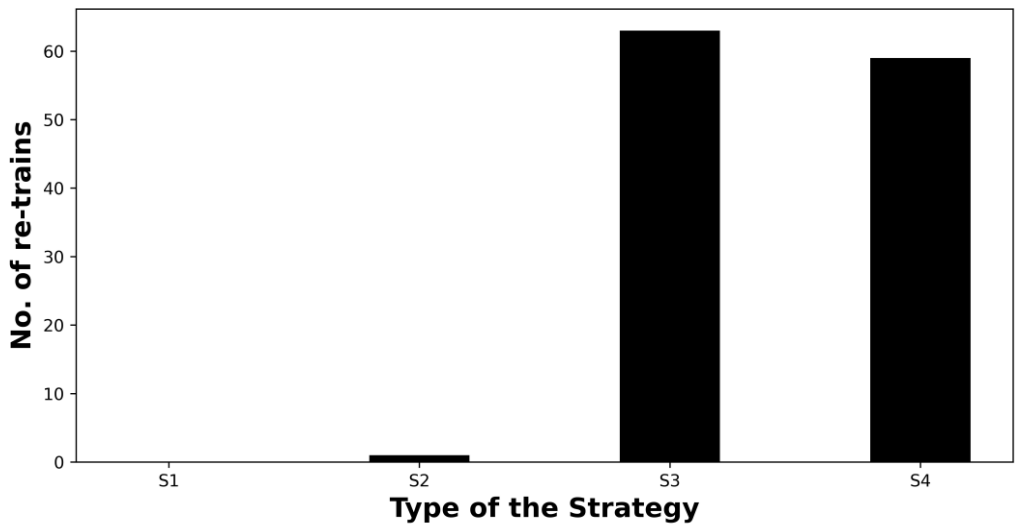

**Fig. 2. Number of retrains invoked for each strategy on churn prediction dataset***
*where S1: Without ND measures; S2: ROSFD + DDM; S3:ROSFD + EDDM; S4: ROSFD + ADWM

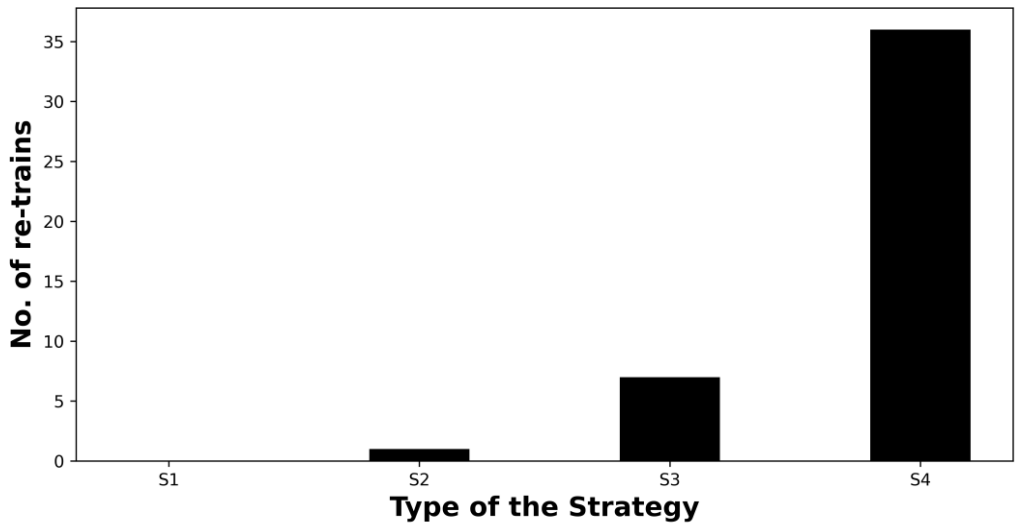

**Fig. 3. Number of retrains invoked for each strategy on credit card prediction dataset**
*where S1: Without ND measures; S2: ROSFD + DDM; S3: ROSFD + EDDM; S4: ROSFD + ADWM

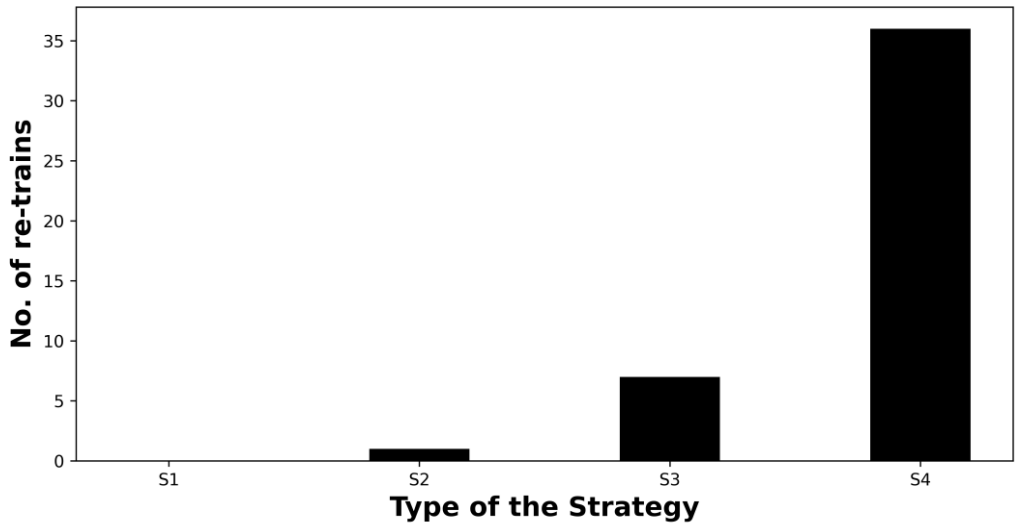

**Fig. 4. Number of retrains invoked for each strategy on ethereum fraud detection dataset**
*where S1: Without ND measures; S2: ROSFD + DDM; S3: ROSFD + EDDM; S4: ROSFD + ADWM

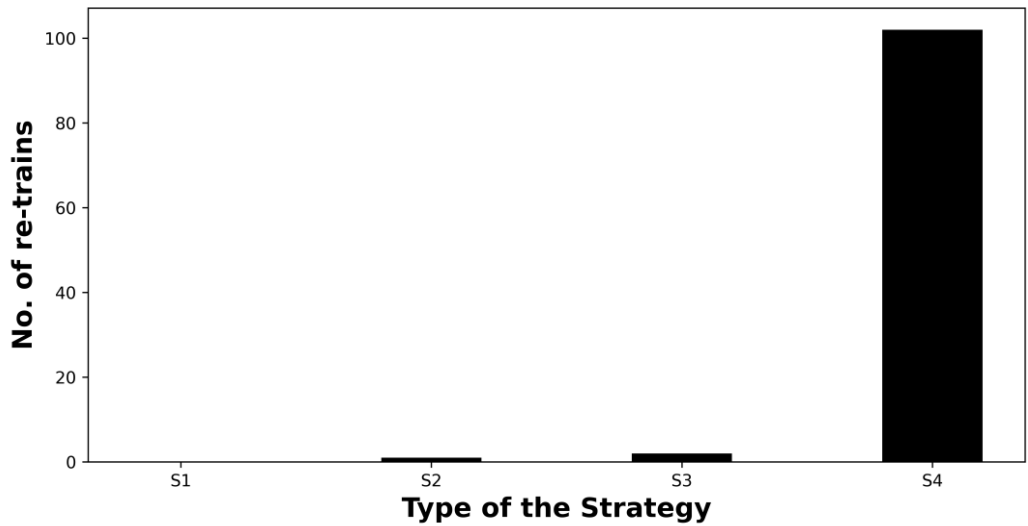

Fig. 5. Number of retrains invoked for each strategy on DDoS on SDN dataset
*where S1: Without ND measures; S2: ROSFD + DDM; S3:ROSFD + EDDM; S4: ROSFD + ADWM

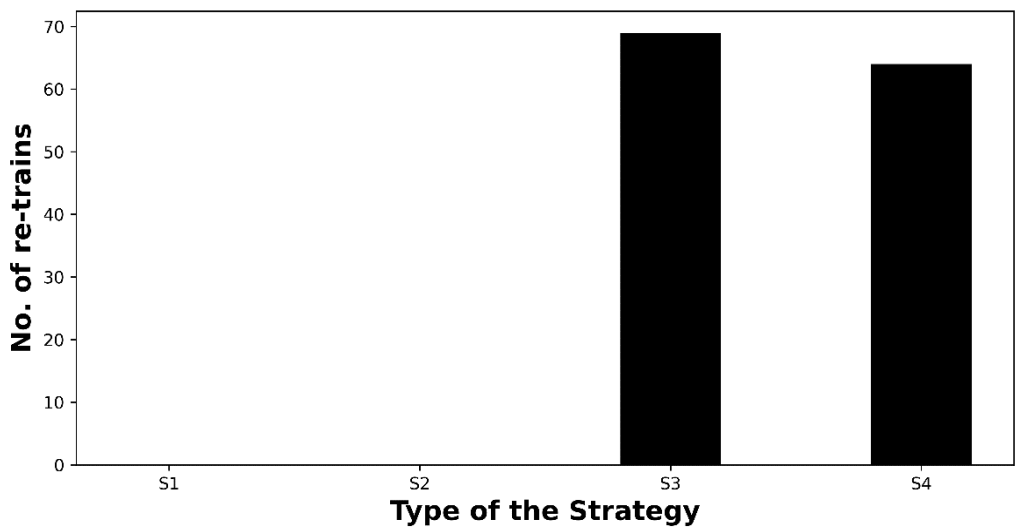

Fig. 6. Number of retrains invoked for each strategy on phishing detection dataset
*where S1: Without ND measures; S2: ROSFD + DDM; S3:ROSFD + EDDM; S4: ROSFD + ADWM

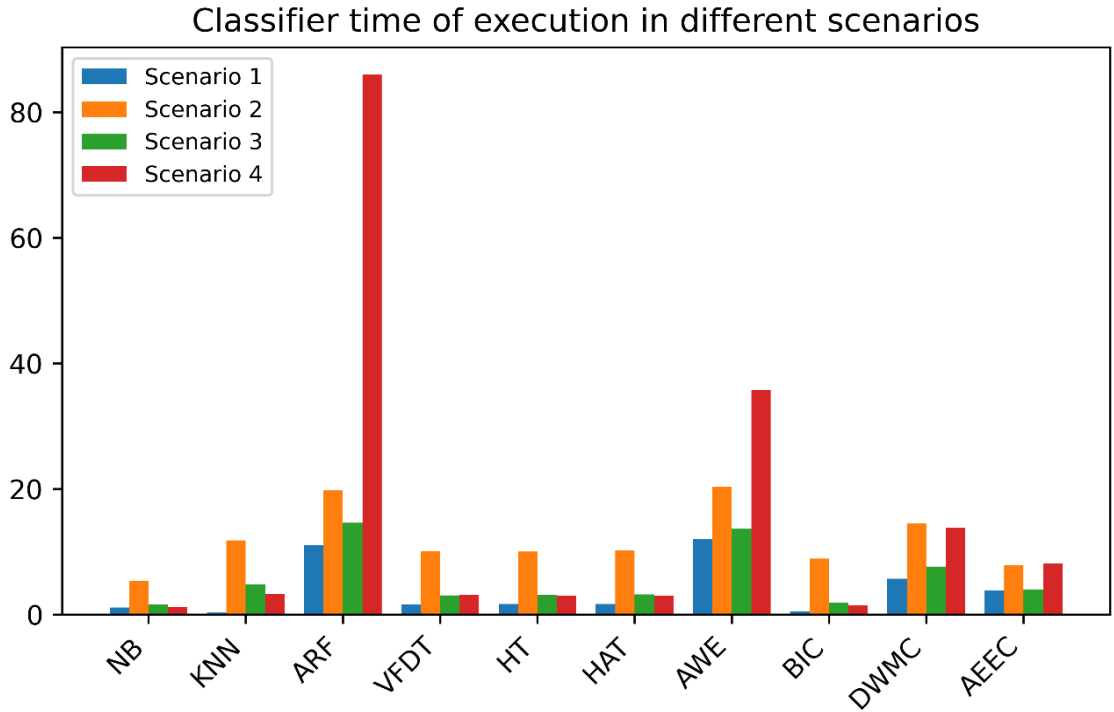

**Fig. 7.** Execution time of various classifiers on churn prediction dataset

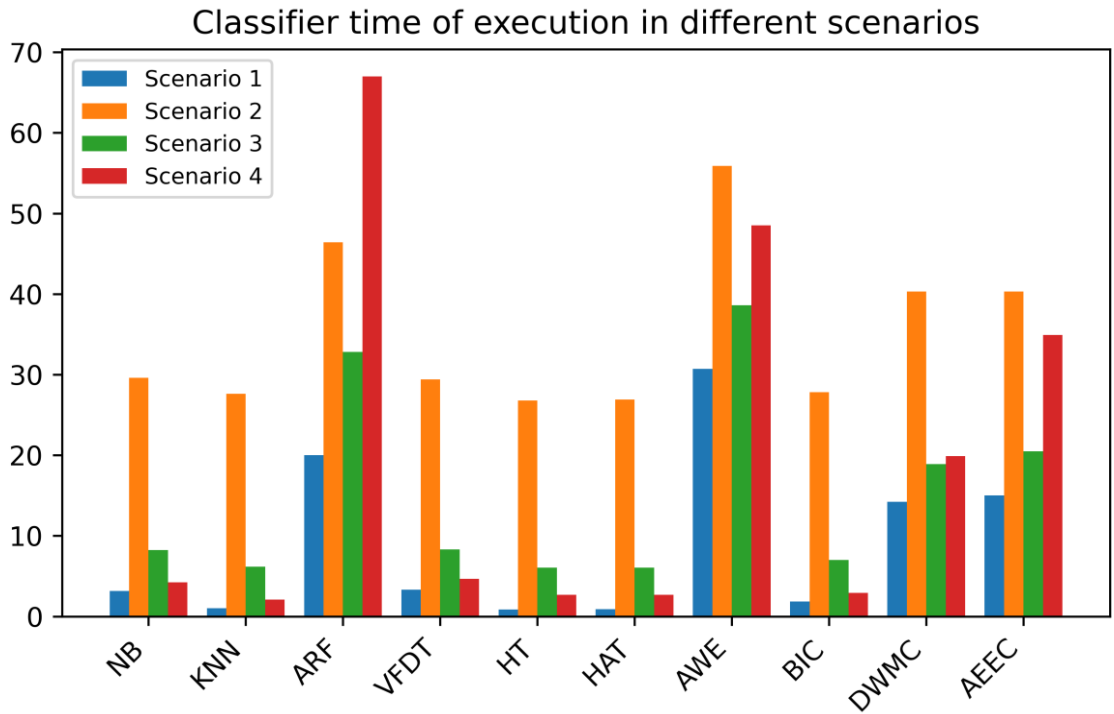

**Fig. 8.** Execution time of various classifiers on credit card default prediction dataset

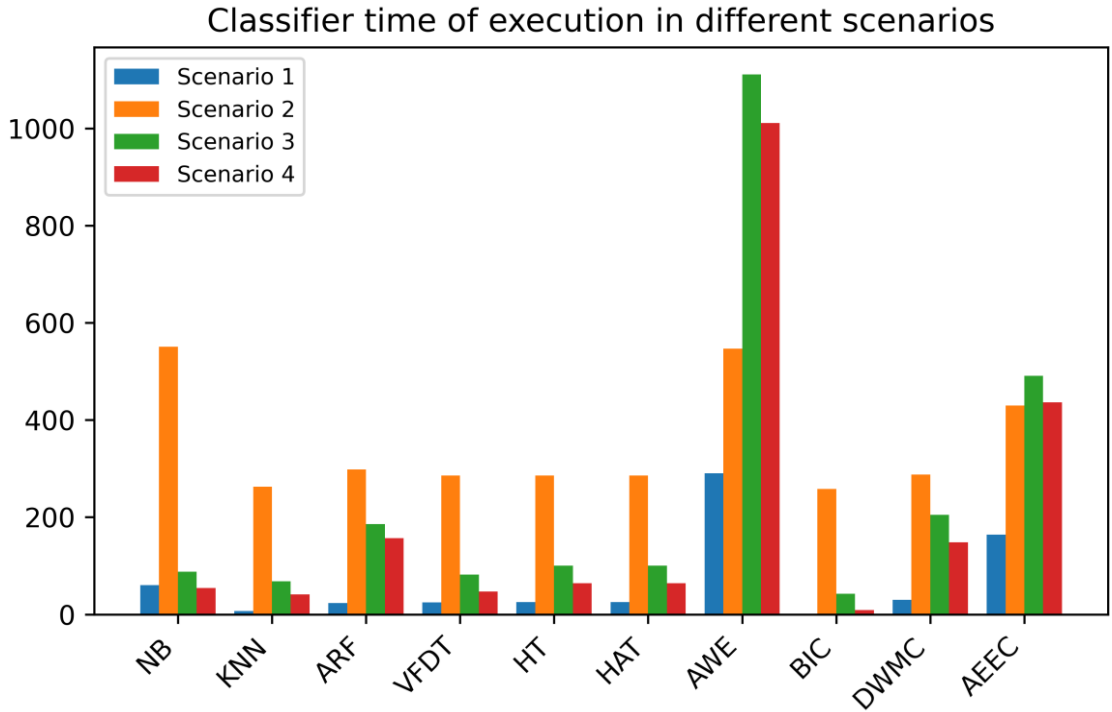

**Fig. 9. Execution time of various classifiers on ethereum fraud detection dataset**

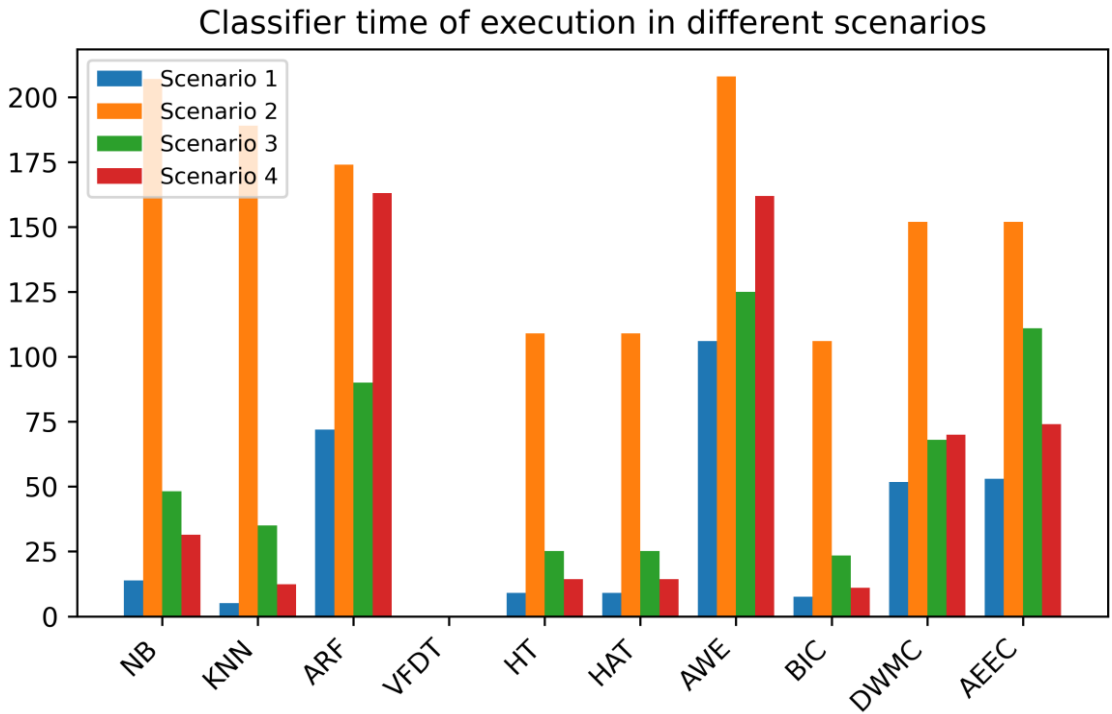

**Fig. 10. Execution time of various classifiers on DDoS on SDN dataset**

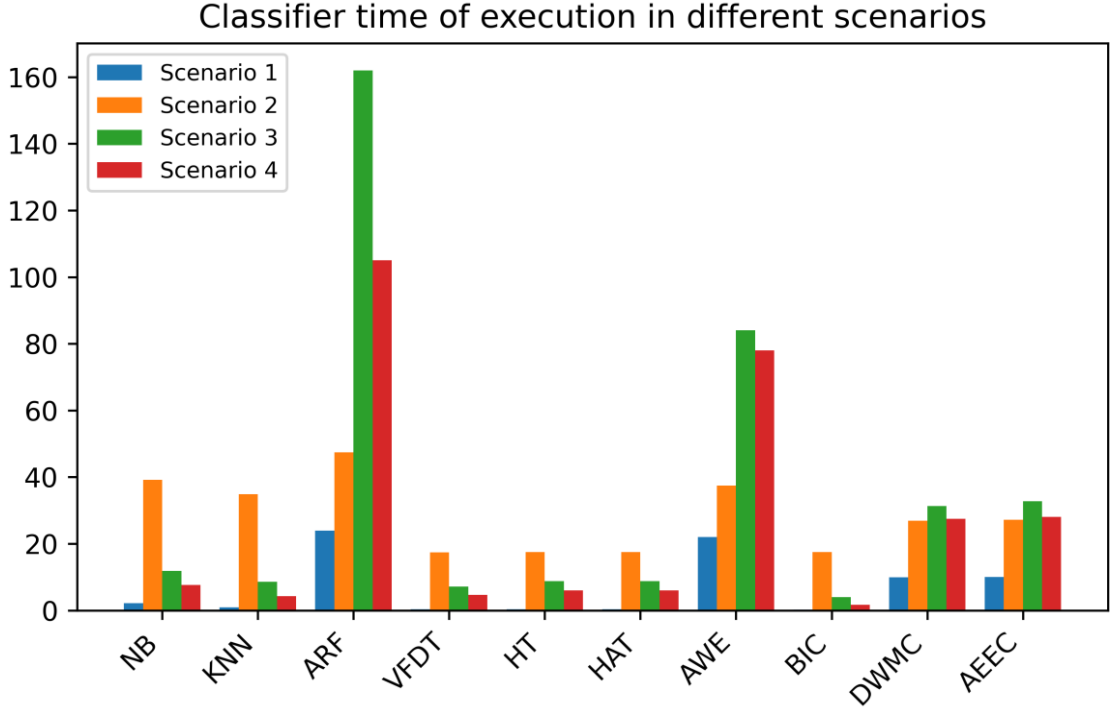

**Fig. 11. Execution time of various classifiers on phishing detection dataset**

## 5. Conclusions

In this paper, we propose ROSFD, a two-stage framework to mitigate the challenges posed by concept drift through the integration of diverse novelty detection methods, specifically DDM, EDDM, and ADWM. Our analysis, conducted across five distinct datasets encompassing the financial and cybersecurity domains, demonstrably indicates that ROSFD, when employing ADWM as the ND metric, consistently achieved the most robust performance in adapting to evolving data patterns. It is important to note that the ARF model emerged as the overall best-performing classifier across the evaluated scenarios. Further, we conducted a comprehensive cost-benefit analysis, meticulously considering both the computational execution time and the number of model retrains, to provide a holistic assessment of the proposed framework's practical utility and efficiency in real-world streaming environments.

## Appendix

### A.1 Overview of the employed ML models

#### A.1.1 Naive Bayes

Naive bayes is probabilistic model based on Bayes theorem [34] and is widely used for many classification tasks. These classifiers are computationally efficient and easy to implement. These models rely on two key assumptions: (i) feature independence and (ii) equal feature contribution to the target class. These assumptions often restrict their practical effectiveness and their applicability in real world.

### A.1.2 K Nearest Neighbours

K Nearest Neighbour (KNN) [35] is a non-parametric supervised learning algorithm. It predicts the target class by utilizing the similarity of the incoming sample with its corresponding neighbours in the training dataset.

### A.1.3 Adaptive Random Forest

Adaptive Random Forest (ARF) [36] is an advanced variant of Random Forest for streaming data and dynamic environment. It dynamically replaces outdated branches / trees with new trees trained on the recent data. The three most important aspects of Adaptive Random Forest are as follows: (i) diversity inclusion via re-sampling; (ii) diversity inclusion via randomly selecting subsets of features; and (iii) drift detectors are placed in each tree and based on the drift, that specific tree is updated accordingly.

### A.1.4 Hoeffding Tree

Hoeffding tree [38] is an incremental decision tree algorithm that is specifically designed for streaming applications. It employs Hoeffding bound, statistical inequality measure at the leaf nodes, and the splitting of the nodes is done only after the difference between top two attributes exceeds a certain threshold. This is employed to improve the probability of the decisions made after splitting nodes.

### A.1.5 Very Fast Decision Tree

Very fast decision rules (VFDR) [37] is an incremental rule learning classifier. The learning mechanism of VFDR and Hoeffding tree is similar, but the key difference is as follows: (i) In the Hoeffding tree, the collected is collected done through the tree, and (ii) whereas, VFDR, this is achieved through rules. Each rule is a combination of conditions that is based on variable values.

### A.1.6 Hoeffding Adaptive Tree

Hoeffding adaptive tree [39] also works in a similar way to the hoeffding tree as discussed in Section A.1.4. The key difference between the hoeffding tree and hoeffding adaptive tree is that the former model assumes that the data distribution is stationary and does not adapt to changes. However, the latter algorithm employs concept drift by detecting changes in data distribution and then updates the model.

### A.1.7 Accuracy Weighted Ensemble

Accuracy weighted ensemble [40] classifier is specifically designed for handling concept drift in data streams. This is used to address the challenges for classifying samples with very large data streams where the underlying data distribution may change over time. It adopts weighting based mechanism, where in the ensemble is assigned a weight based on its expected accuracy on the current data distribution.

### A.1.8 Dynamic Weighted Majority Classifier

Dynamic weighted majority classifier [41] is also an incremental classifier, that employs four-part strategy to handle concept in evolving data streams: (i) it continuously train base models incrementally, (ii) it dynamically updates model weights based on their prediction accuracy; (iii) it employs expert removal, a mechanism to underperform models to maintain ensemble efficiency; and (iv) it employs expert addition to introduce fresh models and adapt to new patterns.

### A.1.9 Adaptive Expert Ensemble Classifier

Additive Expert Ensemble [42] is an adaptive method for online learning in dynamic environments. It manages a group of models (experts) by continuously adding new ones and pruning underperforming or outdated models. While removing the oldest models ensures predictable error limits, eliminating the weakest performers typically delivers better real-world results. This balance of theoretical guarantees and practical adaptability makes AddExp effective for tasks requiring responsiveness to changing data patterns.